\documentclass{article}

 \usepackage[dblblindworkshop, final]{neurips_2026}
\workshoptitle{Foundation Models for Temporal Systems (FMTS)}

\usepackage[utf8]{inputenc} 
\usepackage[T1]{fontenc}    
\usepackage{hyperref}       
\usepackage{url}            
\usepackage{array}
\usepackage{booktabs}       
\usepackage{amsfonts}       
\usepackage{nicefrac}       
\usepackage{microtype}      
\usepackage{xcolor}         
\usepackage{caption}        
\usepackage{amsmath}
\usepackage{graphicx}

\newcommand{\FairAMO}{FairAMO}

\title{Rethinking Agentic Model Orchestration for Time Series Forecasting}

\author{Tong Guan$^{1,2}$\quad Tianxiang Zhan$^{1}$\quad 
 \textbf{Shirui Pan$^{1}$\thanks{Correspondence to: S. Pan $<$s.pan@griffith.edu.au$>$ and M. Jin $<$mingjinedu@gmail.com$>$ } \quad Ming Jin$^{1}$\footnotemark[1]} \\
  $^{1}$Griffith University \quad
  $^{2}$Zhejiang University
}

\begin{document}

\maketitle

\begin{abstract}
Time series foundation models (TSFMs) perform unevenly across forecasting tasks. Recent work uses agentic model orchestration (AMO) to select or combine TSFMs, often outperforming individual forecasters. Yet these gains may reflect stronger or more complementary candidates rather than better orchestration. To disentangle these effects, we introduce \FairAMO, a controlled evaluation protocol with frozen candidate forecasts. Experiments on GIFT-Eval show that {AMO improves accuracy, but its gains are bounded by the candidate TSFMs.} The best orchestrators reduce MASE by less than 6\% relative to the best single TSFM across all tasks. {{These gains depend on complementarity (different candidates excel on different tasks) rather than the number of candidates.}} Adding members yields diminishing returns, and a redundant member can erase these gains. {The value of historical evidence also saturates quickly.} A few recent windows capture all gains, and further history adds none. These findings separate system-level forecasting gains from the contribution of orchestration itself. 
\end{abstract}

\section{Introduction}
Time series forecasting (TSF) supports critical decisions across a wide range of domains~\citep{10.1145/3821637,guan2023spatial,guan2025graphstage,guan2026timeomni}. Recent advances have expanded the forecasting paradigm from task-specific statistical and deep learning models to pretrained time series foundation models (TSFMs) that exhibit strong zero-shot generalization~\citep{shi2025timemoe,zhou2026eidos,guan2026timeomnivl}. This development has also motivated comprehensive benchmarks such as GIFT-Eval~\citep{aksu2024giftevalbenchmarkgeneraltime}, which covers more than 144,000 time series and 7 domains. Results on such diverse benchmarks show that no single TSFM consistently performs best across all datasets and forecasting settings. TimesFM~\citep{TimesFM} excels in long-horizon forecasting with extended contexts. Moirai~\citep{moirai} robustly handles multivariate inputs with covariates. Chronos~\citep{ansariChronosLearningLanguage2024} captures distinct seasonality in short univariate series. Treating any single model as universally best sacrifices higher accuracy for a significant fraction of real-world inputs. This performance heterogeneity makes model orchestration a promising strategy for practical deployment.
\begin{figure}[h]
    \centering
    \includegraphics[width=\linewidth]{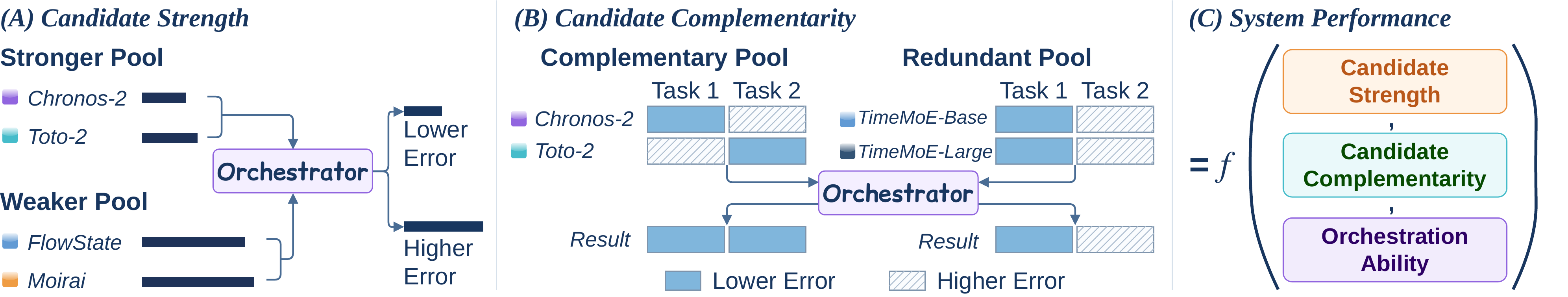}
    \caption{Effects of candidate strength and complementarity on orchestration performance.}
    \label{fig:motivation}
\end{figure}

Recent studies have explored agentic model orchestration (AMO) for selecting or combining candidate TSFMs.
Their reported improvements over individual forecasters are striking~\mbox{\citep{TimeRouter,TSorchestra}}. Notably, AMO systems currently occupy all of the top 10 positions on the GIFT-Eval leaderboard. Yet, leaderboard dominance alone does not establish that these gains originate from the orchestration mechanism, since the systems are evaluated with different candidate pools. This lack of control introduces two fundamental confounding factors. \textbf{\textit{(1) Candidate strength}}, which determines the performance level available to the system before orchestration. As illustrated in Figure~\ref{fig:motivation} (A), the same orchestrator can achieve substantially different results simply because it operates on stronger candidate forecasters. \textbf{\textit{(2) Candidate complementarity}}, which determines the potential headroom available for orchestration. As illustrated in Figure~\ref{fig:motivation} (B), a complementary pool whose members excel on different tasks leaves far more room for orchestration to improve over its best member than a redundant pool whose members succeed and fail in similar settings. Consequently, observed system performance jointly reflects candidate strength, candidate complementarity, and orchestration ability (Figure~\ref{fig:motivation} (C)). Without controlling the first two factors, existing evaluations cannot isolate the contribution of the orchestrator itself.

To disentangle these confounding factors, we introduce \FairAMO, a controlled evaluation protocol built on a Frozen Forecast Bank. Within each comparison, all orchestrators operate on the same candidate forecasts and receive the same explicitly specified historical evidence. Each orchestrator must commit to either a model selection or a set of combination weights before the test labels are revealed, and all resulting forecasts are evaluated using the official GIFT-Eval protocol. By independently varying candidate pool composition, pool size, and evidence budget, our framework separates the contribution of the orchestrator from the properties of the candidate pool. \textbf{Our contributions are threefold.}
(1) We provide a systematic taxonomy and reproducibility analysis of existing AMO systems.
(2) We introduce \FairAMO, a controlled protocol for comparing orchestration methods under identical candidate forecasts, evidence, and evaluation metrics for TSF.
(3) Results show that orchestration gains are strongly conditioned by the candidate pool, while greater orchestration complexity and additional historical evidence do not consistently improve forecasting performance.

\section{Taxonomy and Reproducibility}
\label{sec:tax}
\textbf{Taxonomy.}
We distinguish three adaptive mechanisms according to how orchestration decisions are formed.
(1) \textbf{\textit{LLM-mediated orchestration}} uses an LLM to interpret forecasting evidence and guide model use, as in MoiraiAgent~\citep{MoiraiAgentAgenticFramework}, TSOrchestra's agent~\citep{TSorchestra}, and TimeCopilot's agent framework~\citep{timecopilot}.
(2) \textbf{\textit{Feature-conditioned orchestration}} learns a mapping from time series descriptors to orchestration decisions, as in TimeRouter~\citep{TimeRouter}, Toto-2.0 FnF~\citep{toto_fnf}, and FFORMA~\citep{fforma}.
(3) \textbf{\textit{Historical-error-driven orchestration}} determines decisions directly from past forecasting errors, as in RAES-Conductance-V~\citep{raes_v} and TSOrchestra SLSQP~\citep{TSorchestra}.
The distinction concerns the decision mechanism, not merely whether training uses a loss function; multiple mechanisms may coexist within one system.

\textbf{Reproducibility.}
We examine whether public artifacts support reconstructing an orchestrator under a changed candidate pool and a specified historical evidence budget.
In the releases inspected in September 2026, MoiraiAgent, TimeRouter, and Toto FnF provide inference artifacts, but we did not locate complete training code for their orchestration components; RAES-Conductance-V and TSOrchestra SLSQP expose code for deriving and applying decisions.
Thus, we reproduce AMO methods by reusing or adapting available repository components.

\section{Methodology}
\begin{figure}
  \centering
  \includegraphics[width=1\linewidth]{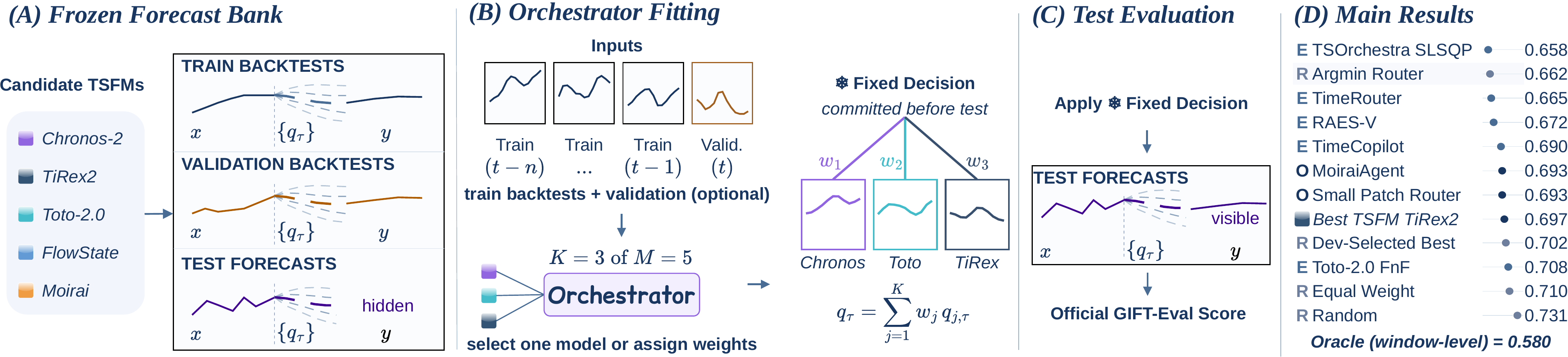}
  \caption{Overview of \FairAMO. (A) Each candidate TSFM is run once and its quantile forecasts are cached as training backtests, validation backtests and test forecasts, with the test labels hidden. (B) The orchestrator reads only the permitted evidence and commits to one model or to a set of weights over $K$ of the $M$ candidates. (C) The frozen decision is applied to the cached test forecasts and scored by the official GIFT-Eval protocol. (D) The 11 deployable orchestrators on the main pool under \texttt{val+B10} (MASE, lower is better); \textcircled{\scriptsize R} , \textcircled{\scriptsize E} , and \textcircled{\scriptsize O}  mark reference, external, and our re-implemented methods. Detailed results are in Appendix Table~\ref{tab:main_pool}.}
  \label{fig:method}
\end{figure}

We treat every orchestrator as an ensemble function of candidate TSFMs. The framework has three stages, illustrated in Figure~\ref{fig:method}. (1) \textbf{\textit{Frozen Forecast Bank.}}
All $M$ candidate TSFMs are run once on GIFT-Eval and are never retrained or fine-tuned afterwards. Their outputs are cached in a forecast bank indexed by \textit{(dataset, term, series, window, candidate, quantile)}, where dataset and term (short, medium, or long horizon) identify a GIFT-Eval task and window a rolling forecast origin. Each entry stores the input history and the candidate's quantile forecasts. The bank has three regions: training backtests and optional validation backtests, both with ground truth attached, and test forecasts, whose ground truth is withheld. Every orchestrator draws its $K \le M$ members from this same bank.
(2) \textbf{\textit{Orchestrator Fitting.}}
Before any test label is revealed, the orchestrator receives the permitted evidence: the $N$ most recent training backtests, plus the validation backtests if available. Using only this evidence, it produces an orchestration strategy, either a model selection or a set of combination weights, which is then committed and frozen for the test period. No rerun, reweighting, or inspection of test scores is permitted after commitment.
(3) \textbf{\textit{Test Evaluation.}}
The frozen strategy is applied to the cached test forecasts to produce the final prediction. Test labels are revealed only at this stage, and the official GIFT-Eval scorer then returns MASE and CRPS.

\textbf{Evaluated Orchestrators.}
We evaluate three groups of orchestrators as shown in Figure~\ref{fig:method} (D). \textbf{\textit{\textcircled{\scriptsize R} Reference orchestrators}} are five simple methods from no learning to perfect hindsight. \emph{Random} samples one member of the candidate subset under a fixed seed. \emph{Dev-Selected Best} uses only the permitted development evidence to nominate a single member applied to every config. \emph{Equal Weight} averages the forecasts of the $K$ members with weight $w_j = 1/K$ assigned to member $j$, without learning any weights. \emph{Argmin Router} commits, per config, to the member with the smallest evidence loss, $\hat{j} = \arg\min_j L_{\text{evidence}}(j, \text{config})$, the official GIFT-Eval loss of member $j$ on its permitted evidence windows. \emph{Oracle} selects the best member per input window from test labels and is reported only as a non-deployable upper bound. \textbf{\textit{\textcircled{\scriptsize E} External orchestrators}} are the published methods introduced in Section~\ref{sec:tax}: TimeRouter~\citep{TimeRouter}, Toto-2.0 FnF~\citep{toto_fnf}, RAES-Conductance-Ensemble-V~\citep{raes_v}, TSOrchestra SLSQP~\citep{TSorchestra}, and TimeCopilot Median~\citep{timecopilot_median}; the training procedures of TimeRouter and Toto-2.0 FnF are not fully public, so we reproduce them as style adaptations. \textbf{\textit{\textcircled{\scriptsize O} Our re-implementations}} are two orchestrators that we rewrite from published designs to fit \FairAMO: \emph{MoiraiAgent-Retrained}~\citep{MoiraiAgentAgenticFramework}, an LLM-mediated orchestrator retrained on the permitted evidence, and \emph{Small Patch Router}, its numeric counterpart, trained on the same pairs and horizon. It patches the history and each candidate's forecast, encodes every candidate with one shared temporal transformer, and scores them by attention across candidates.

\section{Experiments}

\textbf{RQ1. With the candidate pool and the evidence budget held fixed, how much forecasting gain comes from the orchestrator itself rather than from the candidates it selects?}

Existing orchestrators remain far from the ceiling that the candidate pool itself provides, leaving substantial room for improvement. Figure~\ref{fig:method} (D) summarizes the 11 deployable methods in the same 5-candidate setting. The strongest deployable orchestrator, TSOrchestra SLSQP, reaches 0.658 MASE, a gain of 0.039 or 5.6\% over the best single TSFM, TiRex2, at 0.697. Argmin Router reaches 0.662 and TimeRouter 0.665. The window-level oracle sits at 0.580, so even the strongest orchestrator still leaves two thirds of the distance to that ceiling uncovered. Every method that gains more than 0.01 relies on backtest evidence.

\newlength{\FigH}
\newlength{\PanelH}

\begin{figure}[t]
\centering
\setlength{\FigH}{0.23\linewidth}
\begin{minipage}[t]{0.275\linewidth}
  \vspace{0pt}%
  \centering
  \includegraphics[width=\linewidth]{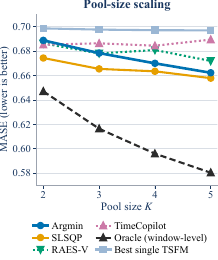}
  \captionof{figure}{Pool-size scaling. The best single TSFM is flat while the oracle falls, so members add complementarity, not strength. Appendix Table~\ref{tab:pool_size}.}
  \label{fig:poolsize}
\end{minipage}\hfill
\begin{minipage}[t]{0.275\linewidth}
  \vspace{0pt}%
  \centering
  \includegraphics[width=\linewidth]{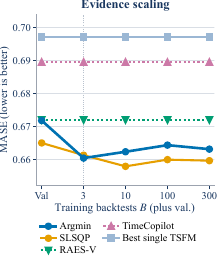}
  \captionof{figure}{Evidence scaling on the main pool. Argmin is best at three backtests and SLSQP at ten; larger budgets bring no further gains. Appendix Table~\ref{tab:evidence}.}
  \label{fig:evidence}
\end{minipage}\hfill
\begin{minipage}[t]{0.425\linewidth}
  \vspace{0pt}%
  \centering
  \includegraphics[width=\linewidth]{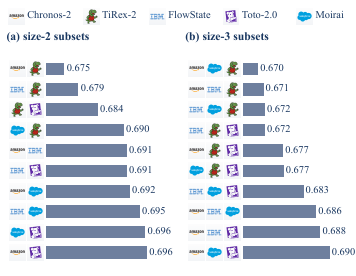}
  \caption{Argmin Router on every size-2 and size-3 candidate subset, sorted by MASE. Brand marks before each bar identify the pool members; the four best pools in each panel all contain TiRex2. Detailed data in Appendix Table~\ref{tab:composition}.}
  \label{fig:pool}
\end{minipage}
\end{figure}

\textbf{RQ2. As the candidate pool grows and as the backtest evidence budget grows, do orchestration gains keep improving, and at what point does each of the two stop paying off?}

On pool size, adding members enlarges the exploitable error diversity, that is the complementarity, rather than the strength of the candidates themselves. Figure~\ref{fig:poolsize} shows the best single TSFM almost unchanged as the pool grows from two to five members, from 0.699 to 0.697, while the window-level oracle drops from 0.647 to 0.580 and the headroom $H$ grows from 0.052 to 0.117. The Argmin Router improves accordingly from 0.689 to 0.662 and its realization rises from 18.4\% to 29.8\%, but the increments shrink: the third member adds 5.1 points of realization, whereas the fourth and the fifth add about 3 points each.

On evidence, a few recent windows carry all of its value, which indicates that orchestration mechanisms depend on the validation segment closest in time to the test windows, and that adding training backtests beyond that point can even hurt. Figure~\ref{fig:evidence} shows the Argmin Router improving from 0.672 with validation only to 0.661 once three training backtests are added, then drifting back to 0.664 as the budget grows to 100 backtests. TSOrchestra SLSQP peaks at 10 backtests with 0.658 and also degrades beyond it, so history past a few recent windows stops helping and can actively hurt.

\textbf{RQ3. At a fixed pool size, how sensitive are orchestration methods to which candidates the pool contains, and what happens to that sensitivity when a weak candidate is introduced?}

Orchestration methods are highly sensitive to composition, no less so than to pool size. Figure~\ref{fig:pool} shows that the same Argmin Router spans 0.670--0.690 MASE over the ten size-3 subsets, a spread of 0.020 that matches in magnitude the 0.027 gained by growing the pool from two to five members, and the spread in realization is wider still, from 9.8\% to 32.3\%, a factor of three. This sensitivity comes from complementarity rather than strength: the best single TSFM of every subset falls between 0.697 and 0.703, a difference of approximately 0.006, whereas subsets containing TiRex2 realize 28.5\% of their headroom on average against 13.4\% for subsets without it.

Introducing a weak candidate does not necessarily hurt; what matters is whether it contributes a new error profile. Toto2-313m is the weakest single model at 0.703, yet adding it to the six size-2 subsets that exclude it improves five of them, by up to 0.013 MASE. Conversely, a large headroom does not guarantee a realizable gain: Chronos2+Toto2-313m has a headroom of 0.048, close to the median, but realizes only 3.1\%, because its members fail on the same configs and the evidence cannot separate them. Pools should therefore be grown by difference in error profile rather than by single-model ranking, and reported orchestration gains must be accompanied by the pool composition, since the realization of one and the same orchestrator can differ by a factor of three.

\section{Conclusion}

Under \FairAMO, the central insight is that AMO improves accuracy, but its gains are bounded by the candidate pool, that {these gains depend on complementarity rather than pool size}, and that {the value of historical evidence saturates quickly}. However, the strongest methods are simple selection and weighting rules, leaving the contribution of LLM-mediated orchestration unclear. A single decision per config also limits realization of the available window-level headroom. Future work may therefore explore finer-grained commitments, agentic methods that use context beyond backtests, and candidate pools built around complementary error profiles rather than model rank.

\bibliographystyle{plainnat}
\bibliography{references}

\appendix
\makeatletter
\setlength{\@fptop}{0pt}   %
\makeatother
\section{Metrics}
\label{app:metrics}

Every number reported in this paper is produced by the official GIFT-Eval scorer on the official test split, so the scores are directly comparable with the public leaderboard. This section defines the two official metrics first and then the quantities we derive from them.

\textbf{Official GIFT-Eval metrics.}
\emph{MASE} is the mean absolute error of the point forecast divided by the in-sample mean absolute error of a seasonal naive forecast on the same series. The division makes errors comparable across series of very different scale, and a value below one means the forecast beats the seasonal naive baseline.
\emph{CRPS} is the continuous ranked probability score of the predicted quantiles, which rewards a well calibrated predictive distribution rather than a single point. The scorer returns both, and GIFT-Eval aggregates each of them over its tasks. Lower is better for both. We report MASE in every table and figure.

\textbf{Units of evaluation.}
A \emph{config} is one GIFT-Eval evaluation setting, that is a dataset paired with a forecast term, either short, medium or long. A \emph{window} is one rolling forecast origin inside a config. Our protocol lets an orchestrator commit one decision per config, and that decision is then held fixed across the windows of the config. This distinguishes fixed per-config decisions from the window-level oracle defined next.

\textbf{Baseline and ceiling.}
The \emph{best single TSFM} of a candidate pool is the member with the lowest MASE over the whole test split, and every gain we report is measured against it. It is deliberately a demanding reference: identifying it requires the test labels, so it is an optimistic stand-in for what a practitioner could actually deploy. \emph{Dev-Selected Best} is its deployable counterpart, nominating one member from the permitted evidence alone. On the main pool it reaches 0.702 against 0.697, and that gap of 0.005 is the cost of having to choose in advance. The \emph{window-level oracle} instead picks, for every window, the candidate with the lowest loss on that window, reading the test labels to do so. It is not deployable and is reported only as a lower bound on what member diversity could in principle deliver.

\textbf{Derived quantities.}
Let $L_{\text{best}}(P)$, $L_{\text{oracle}}(P)$ and $L_{\text{method}}(P)$ be the MASE of the best single TSFM, of the window-level oracle and of an orchestration method on a candidate pool $P$. We report three quantities throughout.

\begin{itemize}
  \item \emph{Headroom} $H(P) = L_{\text{best}}(P) - L_{\text{oracle}}(P)$ is the accuracy that the differences between pool members put on the table. It is a property of the pool alone and does not depend on the orchestrator.
  \item \emph{Gain} $G = L_{\text{best}}(P) - L_{\text{method}}(P)$ is what a method actually recovers relative to the baseline. It is negative for a method that ends up worse than the best single TSFM, which several of the methods we evaluate do.
  \item \emph{Realization} $\text{HR} = G / H(P)$ is the fraction of the available headroom that the method converts into accuracy. By construction HR is $100\%$ for the window-level oracle and negative for a method that loses to the best single TSFM. HR is the quantity that makes pools with different headroom comparable, which is why we use it rather than raw MASE whenever we compare across pools.
\end{itemize}

\textbf{Evidence budgets.}
\texttt{val} denotes the validation backtests on their own. \texttt{val+B}$n$ denotes the validation backtests together with the $n$ most recent training backtests. Unless a table states otherwise, results use the \texttt{val+B10} budget.

\section{Additional Results}
\setcounter{table}{0}
\renewcommand{\thetable}{B\arabic{table}}

This section reports the full results behind the figures in the main text: the main pool in Table~\ref{tab:main_pool}, the effect of pool size in Table~\ref{tab:pool_size}, the effect of the evidence budget in Table~\ref{tab:evidence}, and every size-2 and size-3 subset in Table~\ref{tab:composition}. All metrics are defined in Appendix~\ref{app:metrics}.

\begin{table}[h]
  \centering
  \fontsize{8.5}{10}\selectfont
  \setlength{\tabcolsep}{3pt}
  \renewcommand{\arraystretch}{1.08}
  \caption{Main-pool results under \texttt{val+B10}, grouped as in Figure~\ref{fig:method} (D) into reference, external and re-implemented methods. $G$ is the gain over the best single TSFM and HR is the fraction of the 0.117 headroom realized, so HR $= G/0.117$. Bold marks the best deployable result. The oracle row may switch member per window and is a non-deployable ceiling.}
  \label{tab:main_pool}
  \begin{tabular*}{\linewidth}{@{}>{\raggedright\arraybackslash}p{0.59\linewidth}@{\extracolsep{\fill}}rrr@{}}
    \toprule
    \textbf{Method} & \textbf{MASE} & \textbf{\boldmath$G$} & \textbf{HR (\%)} \\
    \midrule
    \multicolumn{4}{l}{\textbf{Reference}}                               \\
    Oracle (window-level, not deployable) & 0.580 & +0.117   & 100.0   \\
    Argmin Router                         & 0.662 & +0.035   & 29.8    \\
    Dev-Selected Best               & 0.702 & $-$0.004 & $-$3.7  \\
    Equal Weight                          & 0.710 & $-$0.013 & $-$10.9 \\
    Random                                & 0.731 & $-$0.034 & $-$29.2 \\
    \midrule
    \multicolumn{4}{l}{\textbf{External}}                                \\
    TSOrchestra SLSQP~\citep{TSorchestra} & \textbf{0.658} & \textbf{+0.039} & \textbf{33.6}    \\
    TimeRouter~\citep{TimeRouter} & 0.665 & +0.032   & 27.7    \\
    RAES-Conductance-Ensemble-V~\citep{raes_v} & 0.672 & +0.025   & 21.5    \\
    TimeCopilot Median~\citep{timecopilot_median} & 0.690 & +0.008   & 6.5     \\
    Toto-2.0 FnF~\citep{toto_fnf} & 0.708 & $-$0.011 & $-$9.1  \\
    \midrule
    \multicolumn{4}{l}{\textbf{Ours}}                                    \\
    MoiraiAgent-Retrained~\citep{MoiraiAgentAgenticFramework} & 0.693 & +0.004   & 3.8     \\
    Small Patch Router                    & 0.693 & +0.004   & 3.6     \\
    \bottomrule
  \end{tabular*}
\end{table}

\begin{table}[h]
  \centering
  \fontsize{9}{10.5}\selectfont
  \setlength{\tabcolsep}{3pt}
  \renewcommand{\arraystretch}{1.08}
  \caption{Effect of pool size, averaged over all subsets of each size, at the \texttt{val+B10} budget. The ceiling is the window-level oracle and $H$ is the pool headroom, the gap between the best single TSFM and that ceiling. HR is the fraction of $H$ that the Argmin Router realizes. Bold marks the largest headroom and the best Argmin result.}
  \label{tab:pool_size}
  \begin{tabular*}{\linewidth}{@{\extracolsep{\fill}}crrrrr@{}}
    \toprule
    \textbf{Size} & \textbf{Best single TSFM} & \textbf{Ceiling} & \textbf{\boldmath$H$} & \textbf{Argmin MASE} & \textbf{Argmin HR (\%)} \\
    \midrule
    2    & 0.699      & 0.647   & 0.052 & 0.689       & 18.4           \\
    3    & 0.698      & 0.617   & 0.081 & 0.678       & 23.5           \\
    4    & 0.697      & 0.596   & 0.101 & 0.670       & 26.7           \\
    5    & 0.697      & 0.580   & \textbf{0.117} & \textbf{0.662} & \textbf{29.8}           \\
    \bottomrule
  \end{tabular*}
\end{table}

\begin{table}[h]
  \centering
  \fontsize{9}{10.5}\selectfont
  \setlength{\tabcolsep}{3pt}
  \renewcommand{\arraystretch}{1.08}
  \caption{Effect of the evidence budget on the main pool, the sweep plotted in Figure~\ref{fig:evidence}. Entries are MASE (lower is better) and bold marks the best budget for each method whose decision reads the training backtests. The remaining rows are constant because their decisions do not.}
  \label{tab:evidence}
  \begin{tabular*}{\linewidth}{@{}>{\raggedright\arraybackslash}p{0.43\linewidth}@{\extracolsep{\fill}}rrrrr@{}}
    \toprule
    \textbf{Method} & \textbf{Val only} & \textbf{+B3} & \textbf{+B10} & \textbf{+B100} & \textbf{+B300} \\
    \midrule
    Random                        & 0.731 & 0.731 & 0.731 & 0.731 & 0.731 \\
    Equal Weight                  & 0.710 & 0.710 & 0.710 & 0.710 & 0.710 \\
    TimeCopilot Median~\citep{timecopilot_median} & 0.690 & 0.690 & 0.690 & 0.690 & 0.690 \\
    Dev-Selected Best       & 0.703 & \textbf{0.698} & 0.702 & 0.702 & 0.702 \\
    RAES-Conductance-Ensemble-V~\citep{raes_v} & 0.672 & 0.672 & 0.672 & 0.672 & 0.672 \\
    Argmin Router                 & 0.672 & \textbf{0.661} & 0.662 & 0.664 & 0.663 \\
    TSOrchestra SLSQP~\citep{TSorchestra} & 0.665 & 0.661 & \textbf{0.658} & 0.660 & 0.660 \\
    \bottomrule
  \end{tabular*}
\end{table}

\begin{table}[h]
  \centering
  \fontsize{9}{10.5}\selectfont
  \setlength{\tabcolsep}{3pt}
  \renewcommand{\arraystretch}{1.08}
  \caption{The Argmin Router on every size-2 and size-3 subset at the \texttt{val+B10} budget. Best single TSFM is the strongest single member of that subset, $H$ its headroom over the window-level oracle, and HR the fraction of $H$ realized, so HR $=$ (Best single TSFM $-$ MASE)$/H$. Rows are sorted by HR and bold marks the highest within each subset size.}
  \label{tab:composition}
  \begin{tabular*}{\linewidth}{@{}l@{\extracolsep{\fill}}rrrr@{}}
    \toprule
    \textbf{Subset} & \textbf{Best single TSFM} & \textbf{\boldmath$H$} & \textbf{MASE} & \textbf{HR (\%)} \\
    \midrule
    Chronos2+TiRex2               & 0.697 & 0.060 & 0.675 & \textbf{36.4}    \\
    FlowState+TiRex2              & 0.697 & 0.058 & 0.679 & 31.3    \\
    TiRex2+Toto2-313m             & 0.697 & 0.051 & 0.684 & 26.1    \\
    FlowState+Toto2-313m          & 0.702 & 0.055 & 0.691 & 18.7    \\
    Moirai+TiRex2                 & 0.697 & 0.042 & 0.690 & 16.1    \\
    Moirai+Toto2-313m             & 0.703 & 0.047 & 0.696 & 14.9    \\
    FlowState+Moirai              & 0.702 & 0.053 & 0.695 & 13.2    \\
    Chronos2+FlowState            & 0.698 & 0.053 & 0.691 & 12.4    \\
    Chronos2+Moirai               & 0.698 & 0.049 & 0.692 & 12.0    \\
    Chronos2+Toto2-313m           & 0.698 & 0.048 & 0.696 & 3.1     \\
    \midrule
    Chronos2+Moirai+TiRex2        & 0.697 & 0.085 & 0.670 & \textbf{32.3}    \\
    FlowState+TiRex2+Toto2-313m   & 0.697 & 0.082 & 0.672 & 31.1    \\
    Chronos2+FlowState+TiRex2     & 0.697 & 0.085 & 0.671 & 30.5    \\
    FlowState+Moirai+TiRex2       & 0.697 & 0.085 & 0.672 & 29.9    \\
    Moirai+TiRex2+Toto2-313m      & 0.697 & 0.077 & 0.677 & 26.0    \\
    Chronos2+TiRex2+Toto2-313m    & 0.697 & 0.081 & 0.677 & 25.2    \\
    FlowState+Moirai+Toto2-313m   & 0.702 & 0.084 & 0.683 & 22.4    \\
    Chronos2+FlowState+Moirai     & 0.698 & 0.083 & 0.686 & 13.9    \\
    Chronos2+FlowState+Toto2-313m & 0.698 & 0.075 & 0.688 & 13.7    \\
    Chronos2+Moirai+Toto2-313m    & 0.698 & 0.077 & 0.690 & 9.8     \\
    \bottomrule
  \end{tabular*}
\end{table}


\end{document}